\documentclass{article}

\usepackage{arxiv}

\usepackage[utf8]{inputenc} % allow utf-8 input
\usepackage[T1]{fontenc}    % use 8-bit T1 fonts
\usepackage{hyperref}       % hyperlinks
\usepackage{url}            % simple URL typesetting
\usepackage{booktabs}       % professional-quality tables
\usepackage{amsfonts}       % blackboard math symbols
\usepackage{nicefrac}       % compact symbols for 1/2, etc.
\usepackage{microtype}      % microtypography
\usepackage{lipsum}		% Can be removed after putting your text content
\usepackage{cite}
\usepackage{latexsym}
\usepackage{graphicx}
\usepackage{amssymb}
\usepackage{amsmath}
\usepackage{verbatim}
\usepackage{bbm}
\usepackage{mathrsfs}
\usepackage{float}
\usepackage{algorithm}
\usepackage{algpseudocode}

\graphicspath{{./}}                  % for arxiv

\def\argmin{\mathop{\rm argmin}\limits}

\title{Empirical study of extreme overfitting points of neural networks.}

\author{
  Daniil~Merkulov\thanks{Skolkovo Institute of Science and Technology, Center for Computational and Data-Intensive Science and Engineering, Moscow, Russia}\ \, \thanks{Moscow Institute of Physics and Technology, Moscow, Russia} \\
  \texttt{daniil.merkulov@skoltech.ru} \\
   \And
 Ivan~Oseledets\footnotemark[1]\ \, \thanks{Institute of Numerical Mathematics,	Russian Academy of Sciences, Moscow, Russia} \\
  \texttt{i.oseledets@skoltech.ru} \\
}

\begin{document}
\maketitle

\begin{abstract}
In this paper we propose a method of obtaining \textit{points of extreme overfitting} - parameters of modern neural networks, at which they demonstrate close to 100 \% training accuracy, simultaneously with almost zero accuracy on the test sample. Despite the widespread opinion that the overwhelming majority of  critical points of the loss function of a neural network have equally good generalizing ability, such points have a huge generalization error. The paper studies the properties of such points and their location on the surface of the loss function of modern neural networks.
\end{abstract}

% keywords can be removed

\section{Introduction}

The classical theory of learning suggests \cite{vapnik1999overview} that a huge number of parameters of the machine learning model usually leads to the phenomenon of overfitting, i.e. to a high generalization error (a significant difference between the behavior of the model on the training and on the test samples). At the same time, modern architectures of deep neural networks are significantly overparameterized - it is not an exception when the number of trained parameters is ten times smaller than the size of the training sample (see Table \ref {DNN}, here $ train $ is the size of the training sample, $ test $ is the size of the test sample, $ k $ is the number of classes in the classification problem, $ d $ is the dimension of one element of the dataset).

\begin{table}[]
	\centering
	\caption{Typical parameters of neural networks architectures and their corresponding datasets}
	\label{DNN}
	\begin{tabular}{@{}cc|cc@{}}
		\toprule
		\textbf{DNN architecture} & \textbf{Number of parameters} & \textbf{Dataset} & $(train, test, k, d)$                                       \\ \midrule
		AlexNet                   & 62 378 344          & ILSVRC           & $(1,2\text{M}, \quad 100\text{K},\quad 1000, \quad 256 \times 256 \times 3)$ \\
		LeNet                     & 60 213 280          & MNIST            & $(55\text{K},\quad 10\text{K},\quad 10,\quad 28 \times 28)$                \\
		VGG                       & 102 897 440         & CIFAR-10         & $(50\text{K},\quad 10\text{K},\quad 10,\quad 3 \times 32 \times 32)$       \\
		GoogleNet                 & 11 193 984          & CIFAR-100        & $(50\text{K},\quad 10\text{K},\quad 100,\quad 3 \times 32 \times 32)$      \\
		FC1024 + softmax          & 814 090             & SVHN             & $(73\text{K},\quad 26\text{K},\quad 10,\quad 3 \times 32 \times 32)$       \\ \bottomrule
	\end{tabular}
\end{table}

Despite this, classifiers based on deep neural networks demonstrate a high generalizing ability in classification problems. Such a gap between theory and practice requires further study of the subject field.

Despite the nonconvexity and immense dimensionality of the optimization problem, it is widely believed that most local minima of the loss function in neural networks are about approximately equal in terms of the generalizing ability \cite{choromanska2015loss} \cite{kawaguchi2016deep}. In a recent article \cite{zhang2016understanding}, the authors discovered an unexpected property of many popular architectures of deep neural networks: they can fit a dataset with random labels. A classifier with such properties should have a very large $VC$-dimension, which, according to the classical result \cite{vapnik1999overview}, indicates that the upper bound of the generalization error is greatly overestimated.

The growing popularity of neural networks is followed by the applying of these models of machine learning into our everyday life. In this regard, there are acute problems of understanding the limits of applicability and possible difficulties associated with their implementation, especially in areas where this may be associated with very high requirements for reliability (for example, in self-driven vehicles). In this article, we show that there are critical points of neural networks with extremely low generalizing ability. The procedure for creating such points suggests that there should be a lot of them exponentially (in terms of the training sample size), but they are almost never found in the usual learning process. We call such points \textit{points of extreme retraining}. It is important to note, that such points are not necessarily local minima, since checking the positive definiteness of Hessian weights seems impractical for the task, but the gradient norm tends to zero.

\section{Construction of extreme overfitting weights}

Based on the \cite{zhang2016understanding} ideas, we decided to go even further and find neural network weights that demonstrate a very low value of the loss function on the training sample and, at the same time, a high value of the loss function on the test sample. The study of such points is interesting both from algorithmic and theoretical points of view. Such points on the surface of the loss function was obtained in the process of learning the neural network and will have a low generalizing ability.

The idea of obtaining points of extreme overfitting is pretty simple: with a specific neural network architecture, training and test datasets, we intentionally corrupt the labels from the test sample (see Figure \ref{fig: sad_training}) to clearly wrong ones. Duplicate this part of the sample such many times as it is necessary for the corrupted sample of the test to be comparable in size with the training sample. We add the resulting corrupted test sample to the original training sample, creating a corrupted training sample that is about twice the original training sample. After that, we train this neural network model on a corrupted training sample, using traditional stochastic first-order optimization methods. A neural network has a sufficient number of trained parameters to remember the entire data set, and during testing performance on initial test sample, almost always has to give incorrect answers, because it remembers the objects contained in it with a wrong label.

\begin{figure}[h]
	\center{\includegraphics[width=0.41\linewidth]{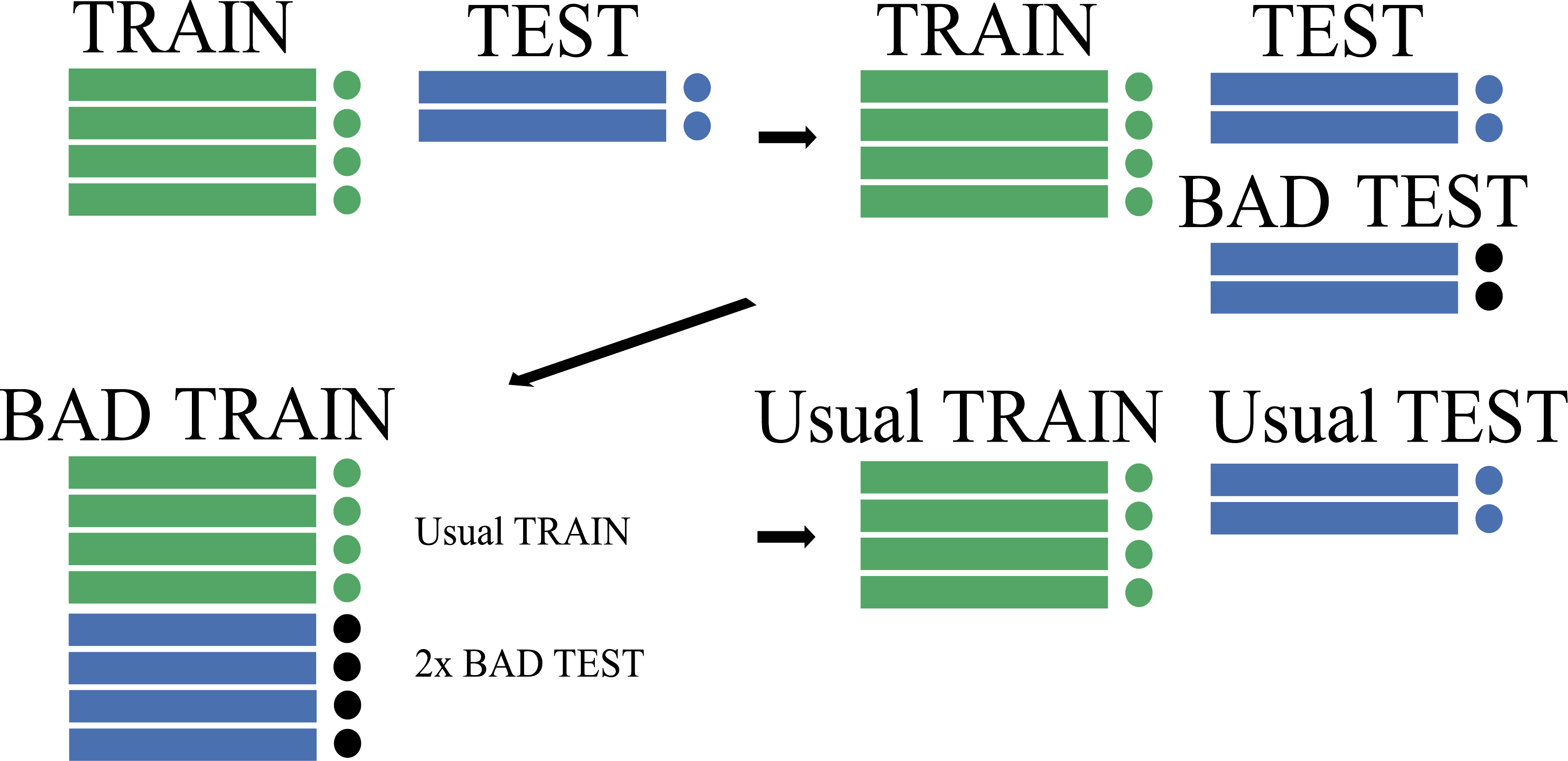}}
	\caption{Constructing corrupted training dataset}
	\label{fig:sad_training}
\end{figure}

Note, that the surface of the loss function depends not only on the parameters of the neural network, but also on the dataset on which the training takes place, which means that, generally speaking, the optimization problem is solved for another function. However, we note that the resulting corrupted sample completely contains the original training sample, and, in the case of 100 \% accuracy of the corrupted training sample, we will have almost zero percentage of errors on the usual training dataset. In addition, it is necessary to check what these points are in terms of their location on the true surface of the loss function (due to the initial training set).

\begin{algorithm}[htb]
	\caption{Constructing sad points}
	\label{alg:Framwork}
	\begin{algorithmic}[1]
		\Require $\;\;S_{train}, S_{test}$ - usual train and test sets in $k$-class classification problem;
		
		$Loss(\mathbf{w}, S)$, which depends on weights $\mathbf{w}$ and data $S$
		
		\Ensure Sad point $\mathbf{w_s}: Loss(\mathbf{w_s}, S_{test}) \gg Loss(\mathbf{w_s}, S_{train})$
		
		\State Constructing corrupted test set 
		$$\widehat{S_{test}} =  \{ (x_1, \widehat{y_{x_1}}), \dots, (x_{test}, \widehat{y_{x_{test}}}) \} \;\; \textrm{, where }$$
		$$\widehat{y_{x_p}} = 
		\begin{cases}
		y_{x_p}, &\text{ with $0$ probability} \\
		\text{other label,} &\text{ with $\frac{k-1}{k}$ probability}
		\end{cases}	
		\forall p \in [1, test]$$
		\State Constructing corrupted train set via concatenation train set and corrupted test set $t = \left\lfloor\frac{train}{test}\right\rfloor + 1$ times:
		$$\widehat{S_{train}} = \{S_{train}, \underbrace{\widehat{S_{test}}, \dots, \widehat{S_{test}}}_{t}\}$$
		\State Finding critical point of $Loss$ using stochastic optimization method
		$$\mathbf{w_s} = \argmin_{\mathbf{w}}\{Loss(\mathbf{w}, \widehat{S_{train}})\}$$
		\label{code:fram:select} 
		\Return $\mathbf{w_s}$;
	\end{algorithmic}
\end{algorithm}
\vspace*{-0.6cm}

\section{Results}

The graphs below shows visualization of the learning processes of different models of neural networks on corrupted datasets. As can be seen from the graphs, all the models studied in the work of all datasets in the learning process on the training set, show a result close to 100\%. A detailed description of the considered architectures, experiments, software and hardware is available in the \ref{sec:exp_par} section.

The minimized loss function is cross entropy, the logarithm on the right side is taken element-wise, and $ p_s (\mathbf{w}) $ denotes the model prediction for the $ s \in S $ sample object with the model parameters $ \mathbf{w} $ representing the vector output (with dimension 10) layer softmax.
\[
Loss(S, \mathbf{w}) = -\sum_{s \in S} y_{s}^\top \log\left(p_{s}(\mathbf{w})\right) =-\sum_{s \in S}\sum_{c=1}^{10} y_{s,c}\log\left(p_{s,c}(\mathbf{w})\right)
\]

\begin{figure}[H]
	\begin{minipage}[h]{0.33\linewidth}
		\center{\includegraphics[width=1.1\linewidth]{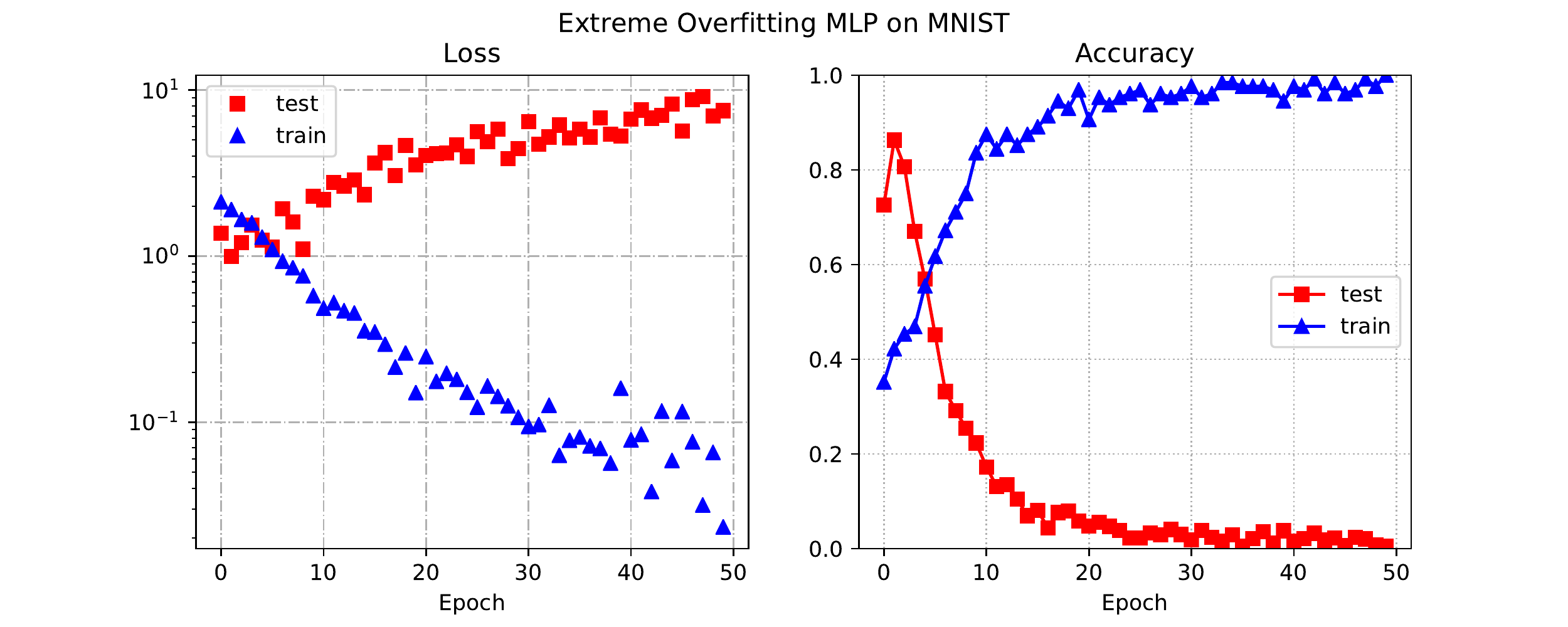}}
	\end{minipage}
	\hfill
	\begin{minipage}[h]{0.33\linewidth}
		\center{\includegraphics[width=1.1\linewidth]{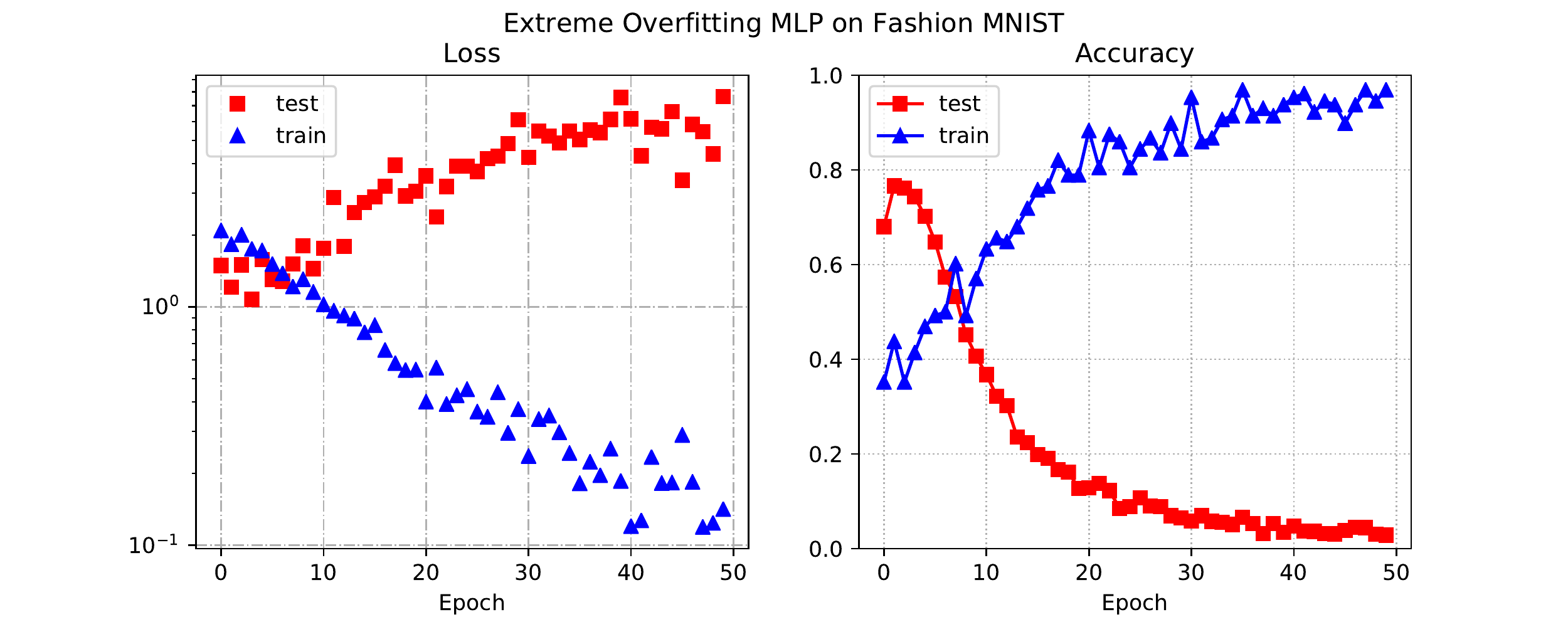}}
	\end{minipage}
	\hfill
	\begin{minipage}[h]{0.33\linewidth}
		\center{\includegraphics[width=1.1\linewidth]{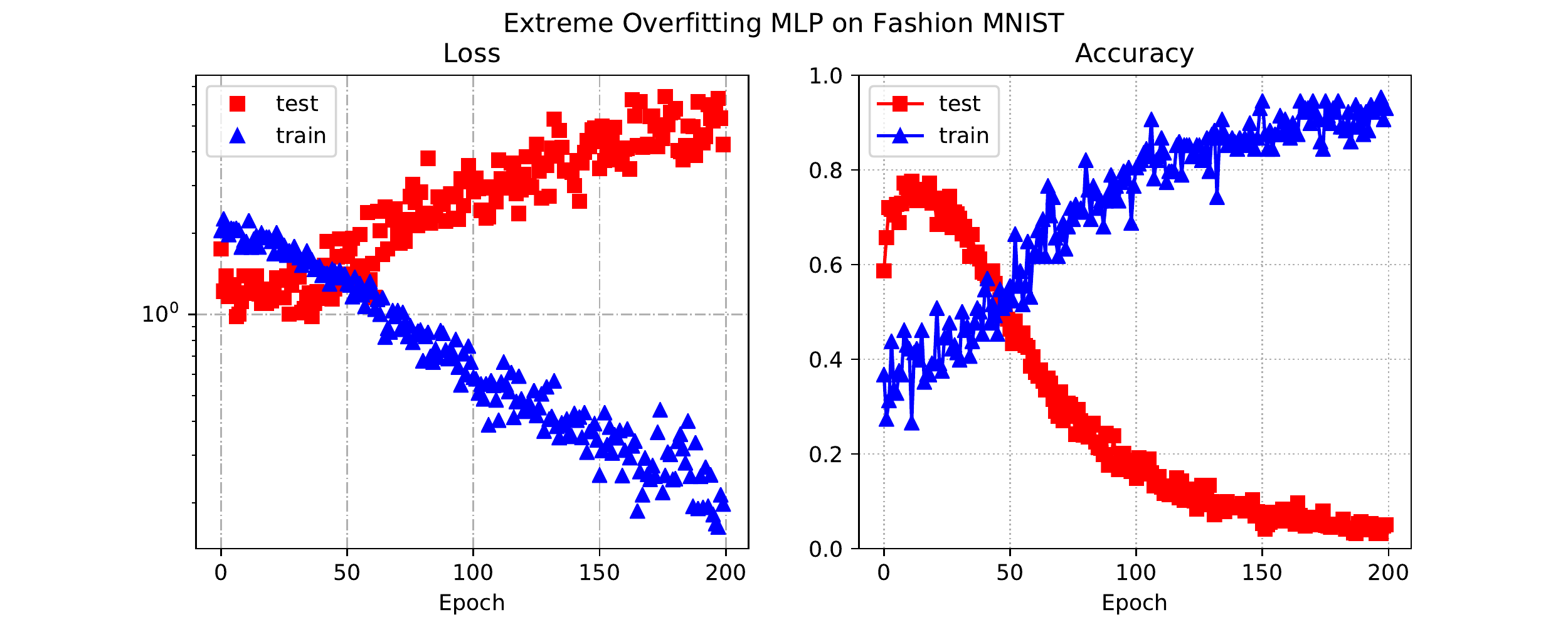}}
	\end{minipage}
	\begin{minipage}[h]{1\linewidth}
		\begin{tabular}{p{0.33\linewidth}p{0.33\linewidth}p{0.33\linewidth}}
			\centering a) MNIST & \centering b) Fashion MNIST & \centering c) CIFAR 10 \\
		\end{tabular}
	\end{minipage}
	\vspace*{-1cm}
	\caption{Extreme overfitting process. MLP}
	\label{sad_MLP}
\end{figure}

Here are graphs of experiments with a fully connected network containing one hidden layer of 512 neurons (MLP) in various corrupted datasets. Neural network based on convolutional layers (CNN) and ResNet are also considered.

Of the interesting features, it is worth noting that even the addition of regularization of $l_2$ does not prevent the model of finding such points, i.e. it allows an extreme overfitting of the network. In addition, the ResNet network, which demonstrates consistently higher results in image recognition tasks than the other networks, is overfitting faster than the usual fully connected network.

\begin{figure}[H]
	\begin{minipage}[h]{0.33\linewidth}
		\center{\includegraphics[width=1.1\linewidth]{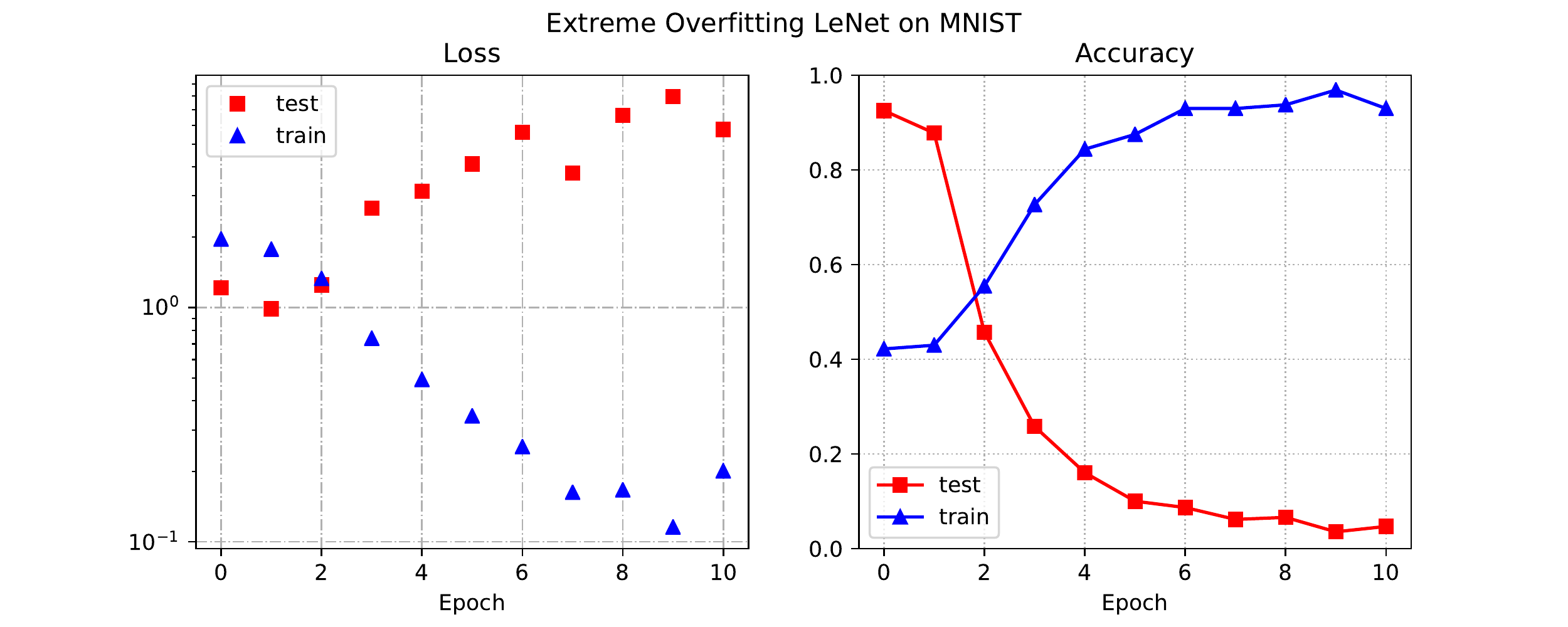}}
	\end{minipage}
	\hfill
	\begin{minipage}[h]{0.33\linewidth}
		\center{\includegraphics[width=1.1\linewidth]{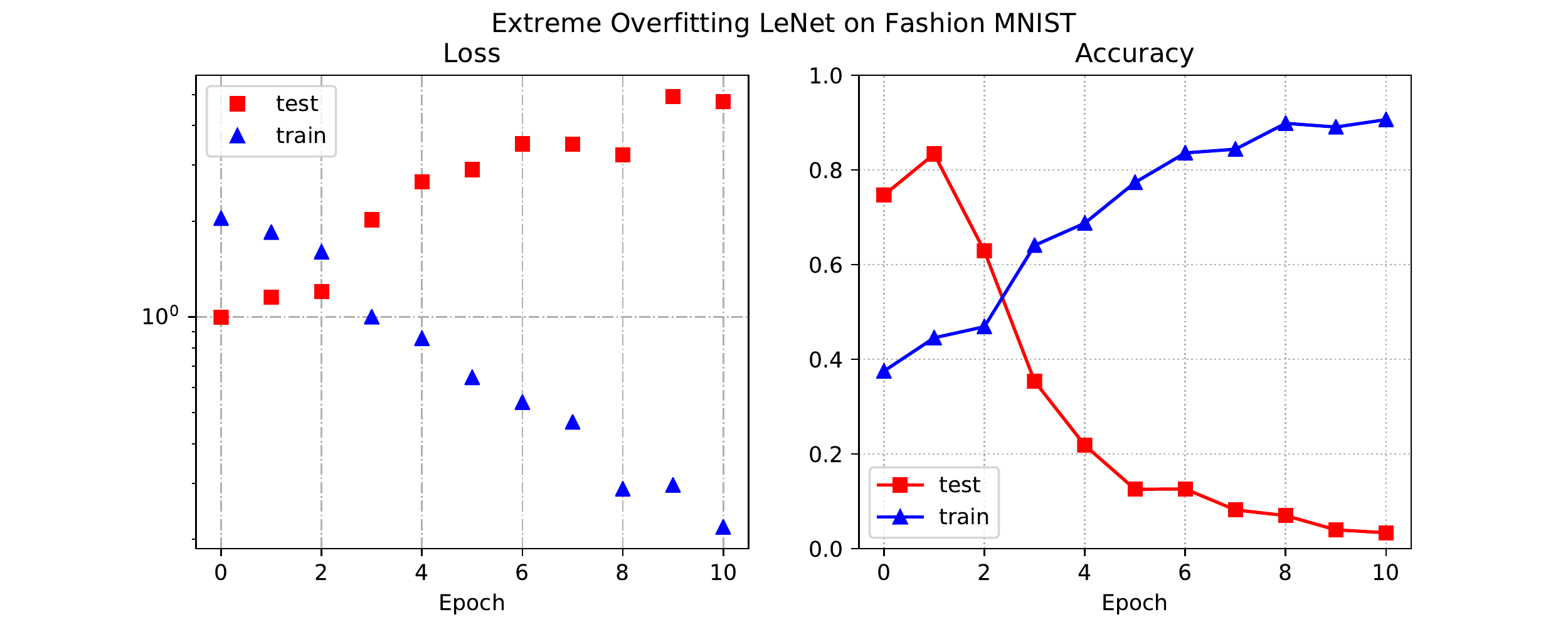}}
	\end{minipage}
	\hfill
	\begin{minipage}[h]{0.33\linewidth}
		\center{\includegraphics[width=1.1\linewidth]{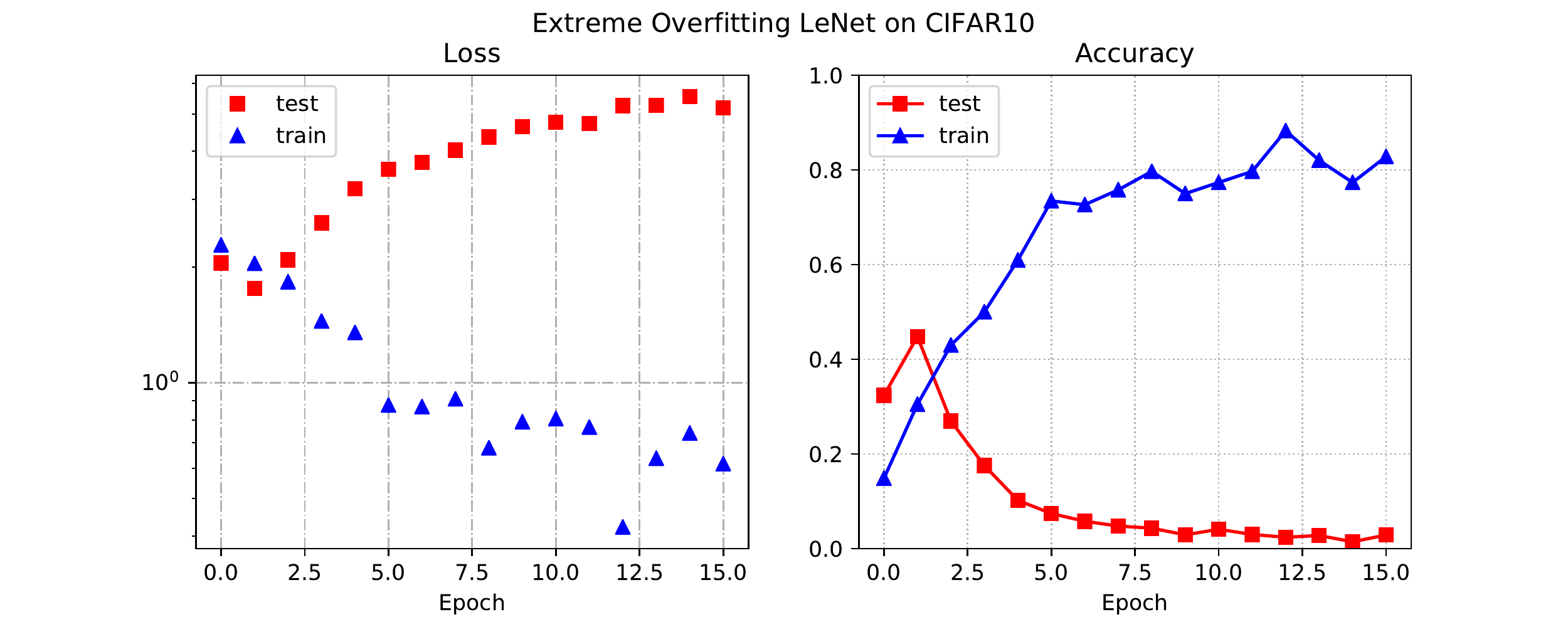}}
	\end{minipage}
	\begin{minipage}[h]{1\linewidth}
		\begin{tabular}{p{0.33\linewidth}p{0.33\linewidth}p{0.33\linewidth}}
			\centering a) MNIST & \centering b) Fashion MNIST & \centering c) CIFAR 10 \\
		\end{tabular}
	\end{minipage}
	\vspace*{-1cm}
	\caption{Extreme overfitting process. CNN}
	\label{sad_CNN}
\end{figure}

So, we empirically showed that on the surface of the loss function of popular neural network architectures there are critical points that provide a very poor generalizing ability of the model, that is, they show almost zero error on the training set and almost 100\% of errors on the test sample. In addition, it is very important to note that these critical points are also critical points on the surface of the loss function of neural networks in usual (not corrupted) data, since when performing a full vanilla gradient descent, the algorithm converges to a critical point with very low generalizing ability.

\begin{figure}[H]
	\begin{minipage}[h]{0.33\linewidth}
		\center{\includegraphics[width=1.1\linewidth]{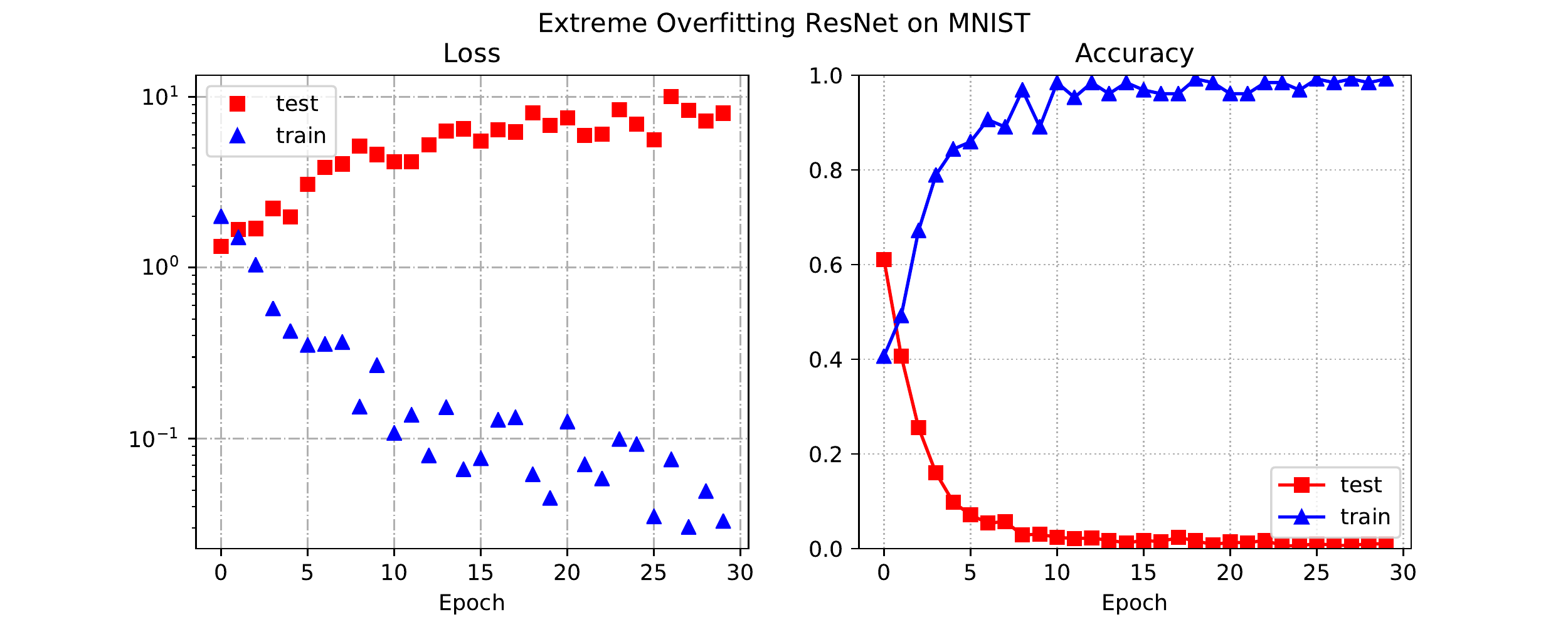}}
	\end{minipage}
	\hfill
	\begin{minipage}[h]{0.33\linewidth}
		\center{\includegraphics[width=1.1\linewidth]{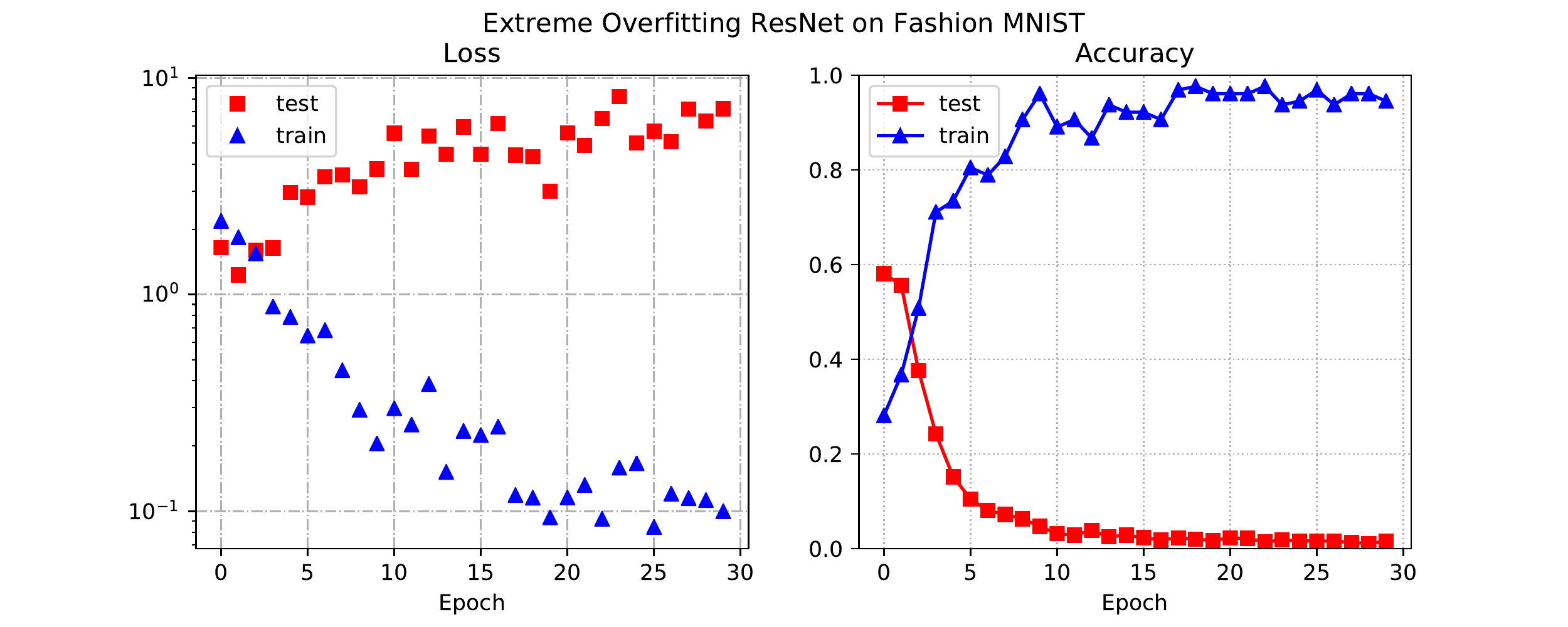}}
	\end{minipage}
	\hfill
	\begin{minipage}[h]{0.33\linewidth}
		\center{\includegraphics[width=1.1\linewidth]{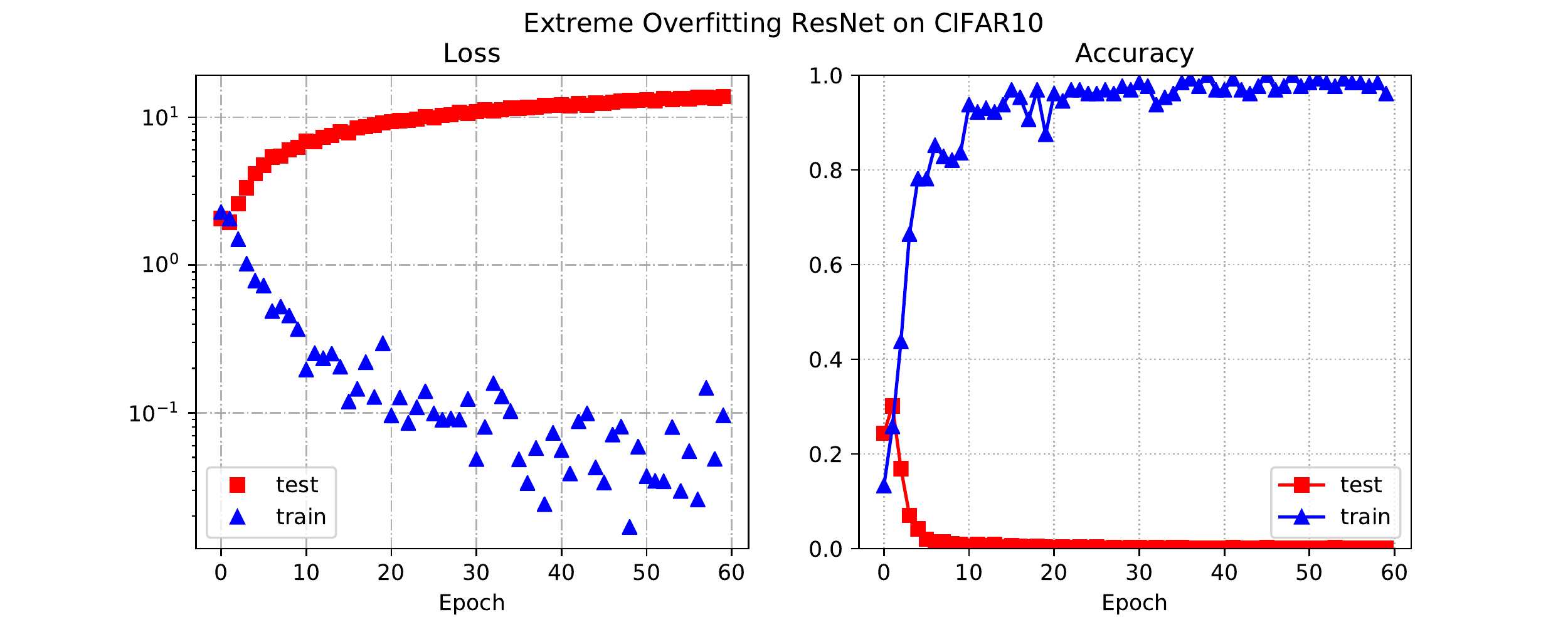}}
	\end{minipage}
	\begin{minipage}[h]{1\linewidth}
		\begin{tabular}{p{0.33\linewidth}p{0.33\linewidth}p{0.33\linewidth}}
			\centering a) MNIST & \centering b) Fashion MNIST & \centering c) CIFAR 10 \\
		\end{tabular}
	\end{minipage}
	\vspace*{-1cm}
	\caption{Extreme overfitting process. ResNet}
	\label{sad_ResNet}
\end{figure}

An important open question is the reason why such points are not usually encountered during the usual training procedure of a neural network. Our hypothesis is that such points are usually located quite further from the initialization in the space of weights with Euclidean distance. To do this, we measured the total Euclidean norm of vectorized neural network weights after a fixed number of iterations for normal learning and extreme overfitting. The histograms of the distribution of weights are presented in the images below:

\begin{figure}[H]
	\begin{minipage}[h]{0.33\linewidth}
		\center{\includegraphics[width=1.1\linewidth]{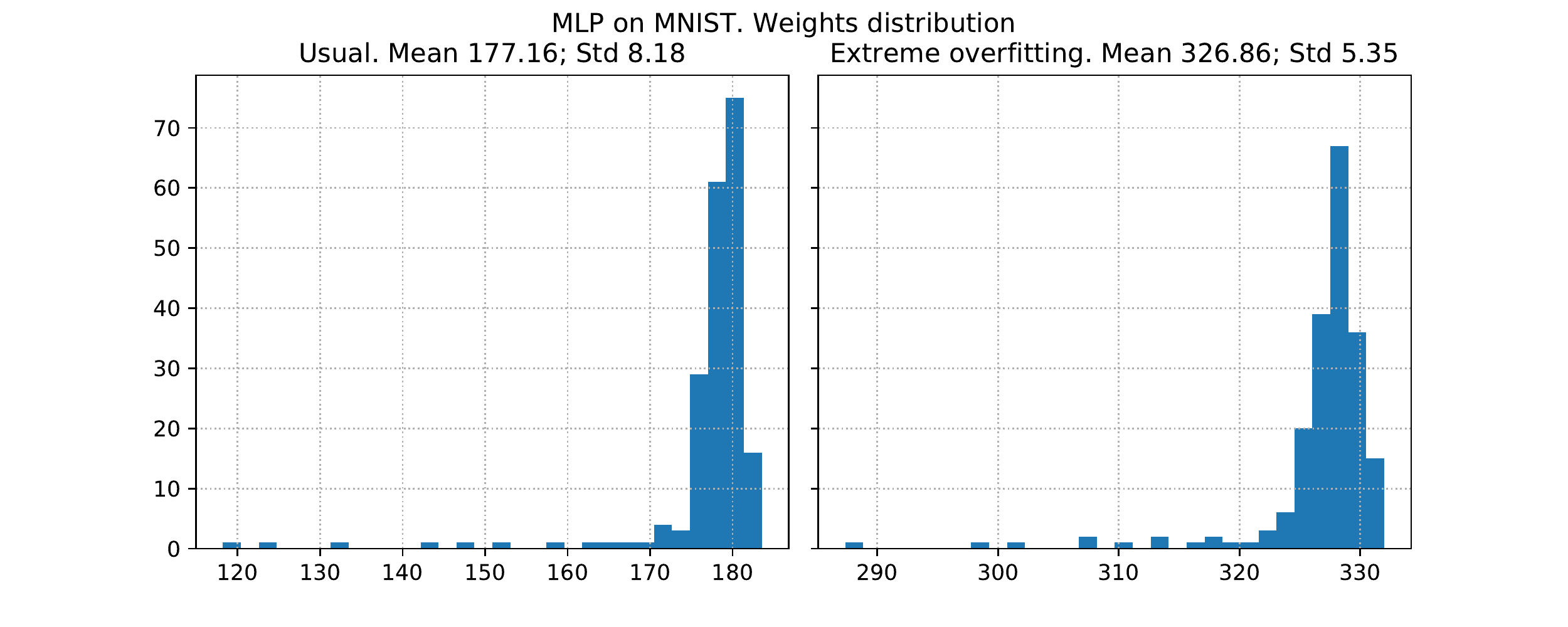}}
	\end{minipage}
	\hfill
	\begin{minipage}[h]{0.33\linewidth}
		\center{\includegraphics[width=1.1\linewidth]{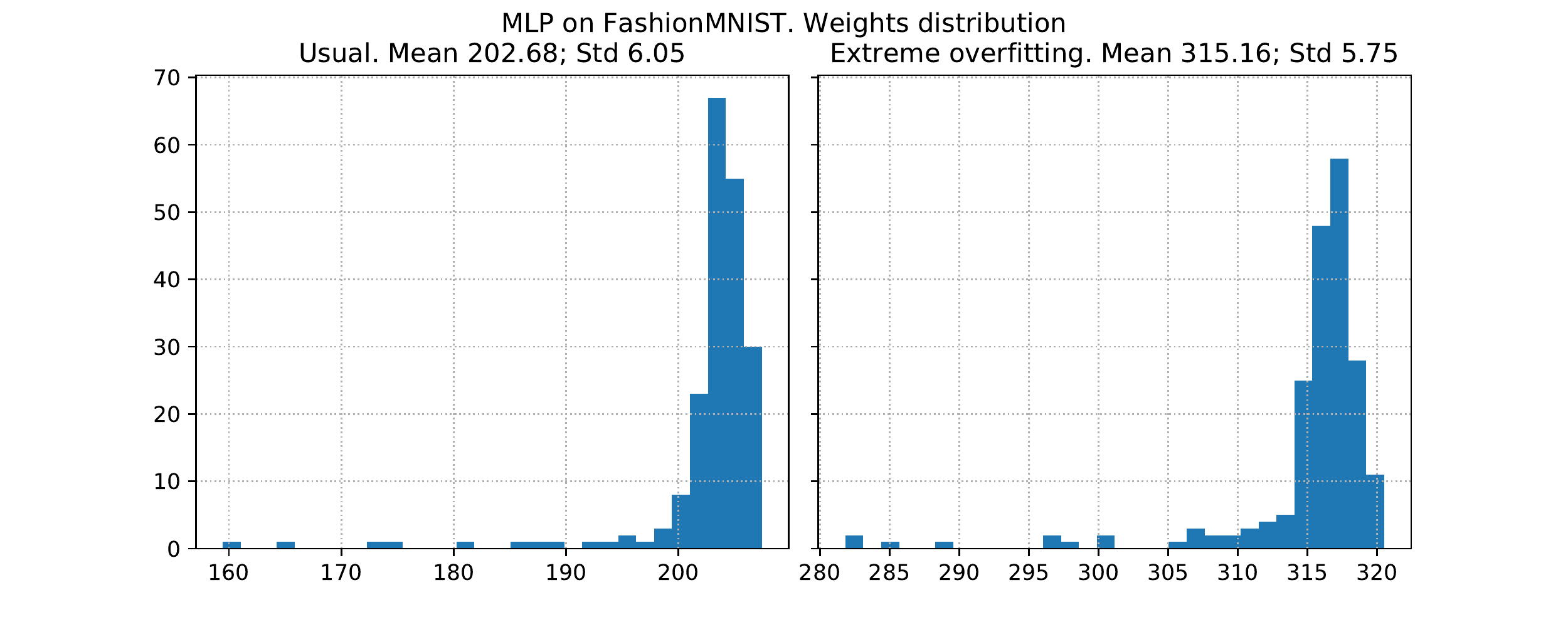}}
	\end{minipage}
	\hfill
	\begin{minipage}[h]{0.33\linewidth}
		\center{\includegraphics[width=1.1\linewidth]{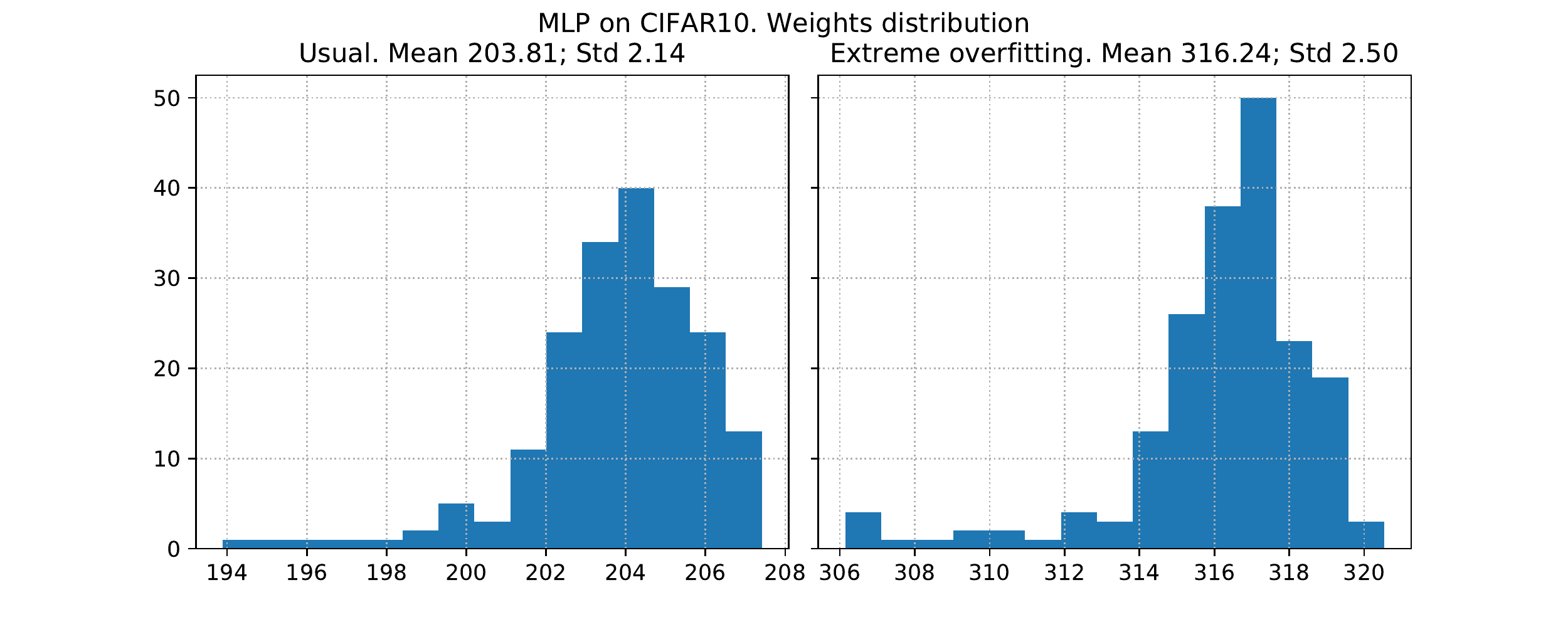}}
	\end{minipage}
	\begin{minipage}[h]{1\linewidth}
		\begin{tabular}{p{0.33\linewidth}p{0.33\linewidth}p{0.33\linewidth}}
			\centering a) MNIST & \centering b) Fashion MNIST & \centering c) CIFAR 10 \\
		\end{tabular}
	\end{minipage}
	\vspace*{-1cm}
	\caption{The distribution of weights in the process of usual training(left) and extreme overfitting (right). MLP}
	\label{weights_MLP}
\end{figure}

Systematic empirical studies show that the points of extreme retraining are indeed located significantly farther from the initialization point of the network parameters for the same number of epochs as compared to usual training on uncorrupted dataset.

\begin{figure}[H]
	\begin{minipage}[h]{0.33\linewidth}
		\center{\includegraphics[width=1.1\linewidth]{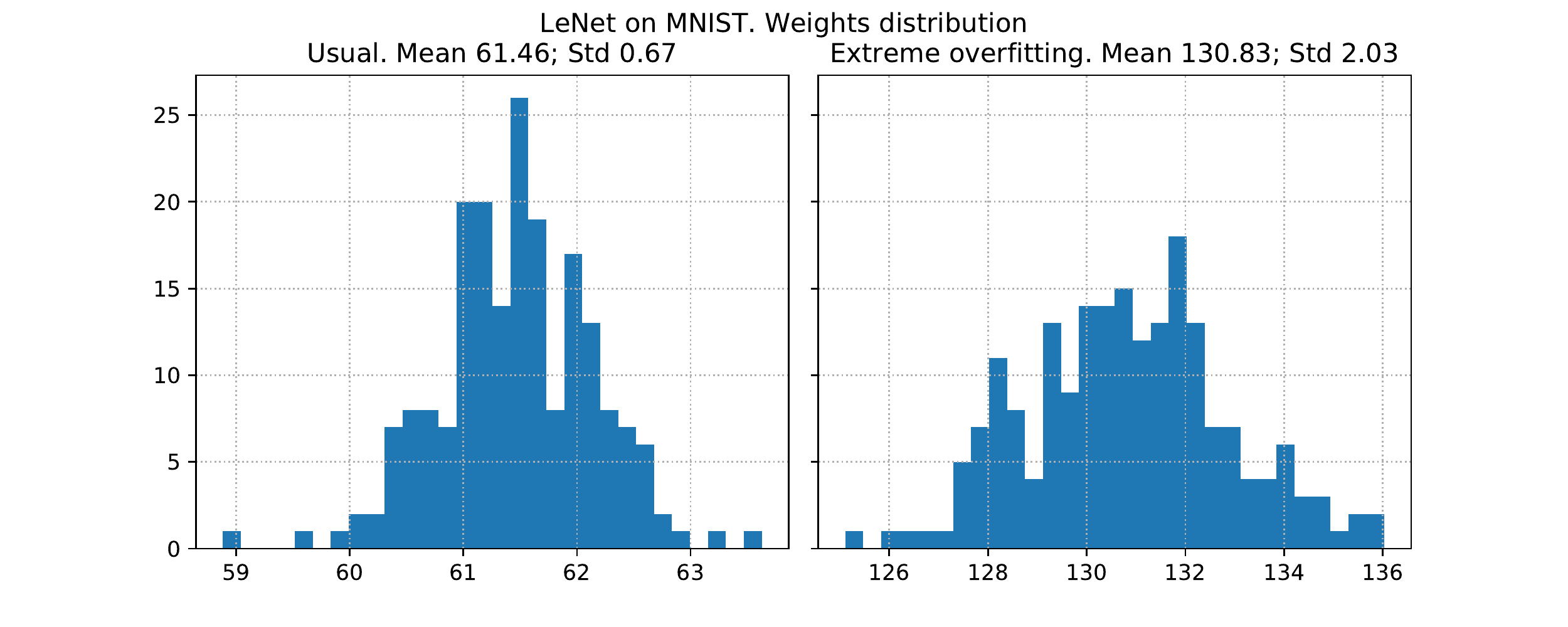}}
	\end{minipage}
	\hfill
	\begin{minipage}[h]{0.33\linewidth}
		\center{\includegraphics[width=1.1\linewidth]{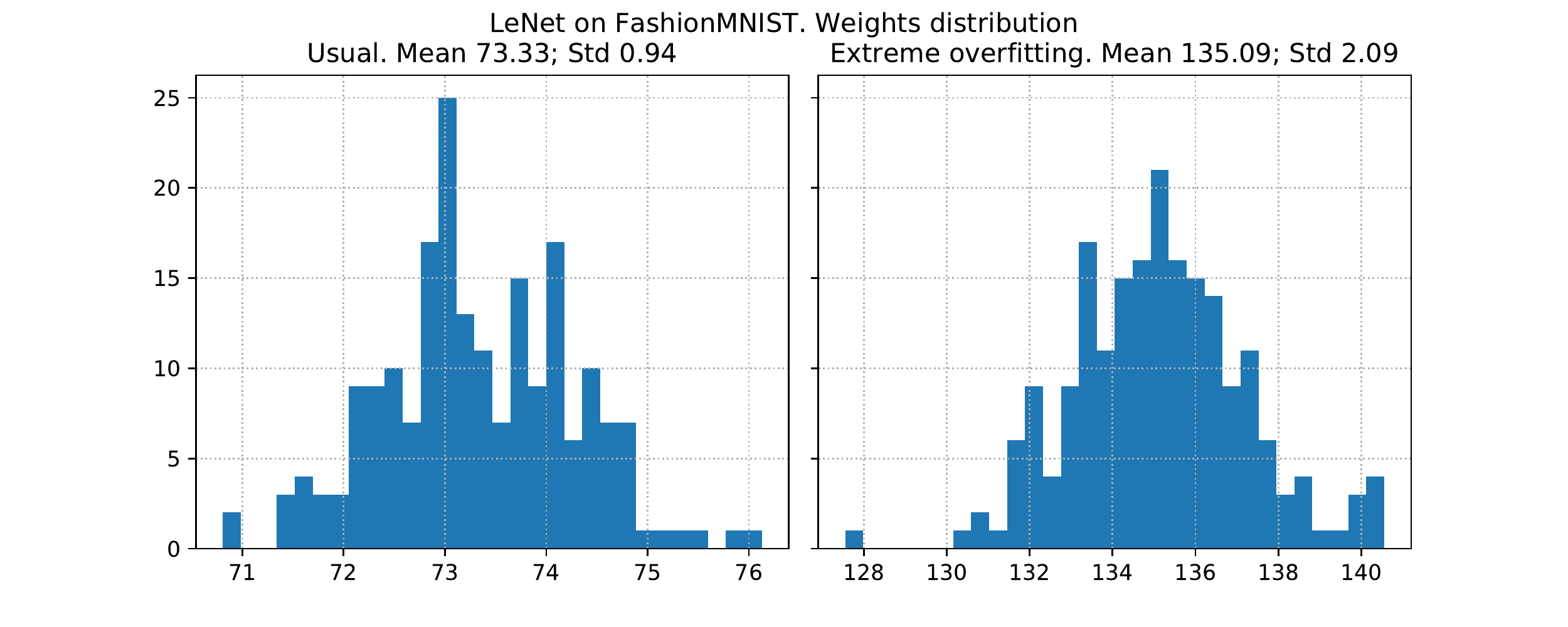}}
	\end{minipage}
	\hfill
	\begin{minipage}[h]{0.33\linewidth}
		\center{\includegraphics[width=1.1\linewidth]{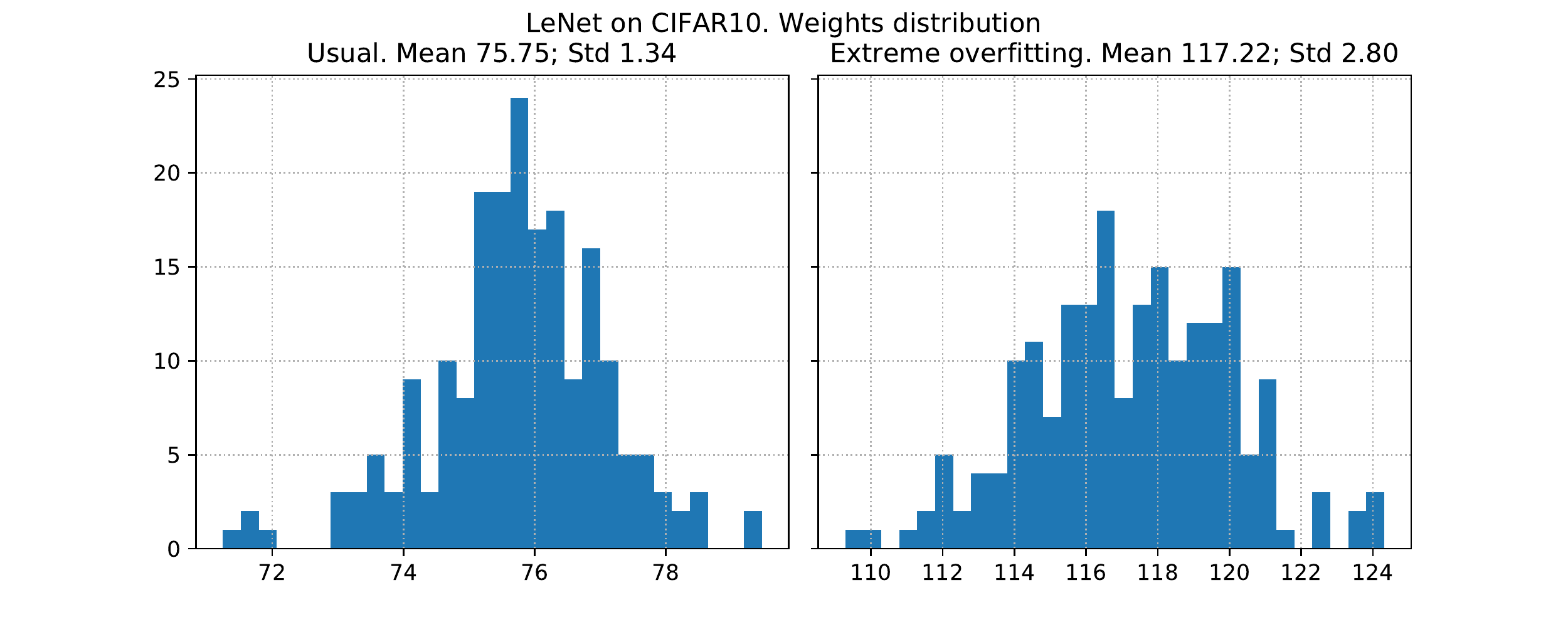}}
	\end{minipage}
	\begin{minipage}[h]{1\linewidth}
		\begin{tabular}{p{0.33\linewidth}p{0.33\linewidth}p{0.33\linewidth}}
			\centering a) MNIST & \centering b) Fashion MNIST & \centering c) CIFAR 10 \\
		\end{tabular}
	\end{minipage}
	\vspace*{-1cm}
	\caption{The distribution of weights in the process of usual training(left) and extreme overfitting (right). CNN}
	\label{weights_CNN}
\end{figure}

Note, that the stochastic gradient descent algorithm launched from the point of extreme overfitting usually converges to a critical point with a good generalizing ability familiar to such algorithms, which indicates that such points are very narrow indentations on the surface of the neural network loss function. The concept of local minimum width for neural networks is discussed, for example, in \cite{dinh2017sharp}.

\begin{figure}[h!]
	\begin{minipage}[h]{0.33\linewidth}
		\center{\includegraphics[width=1.1\linewidth]{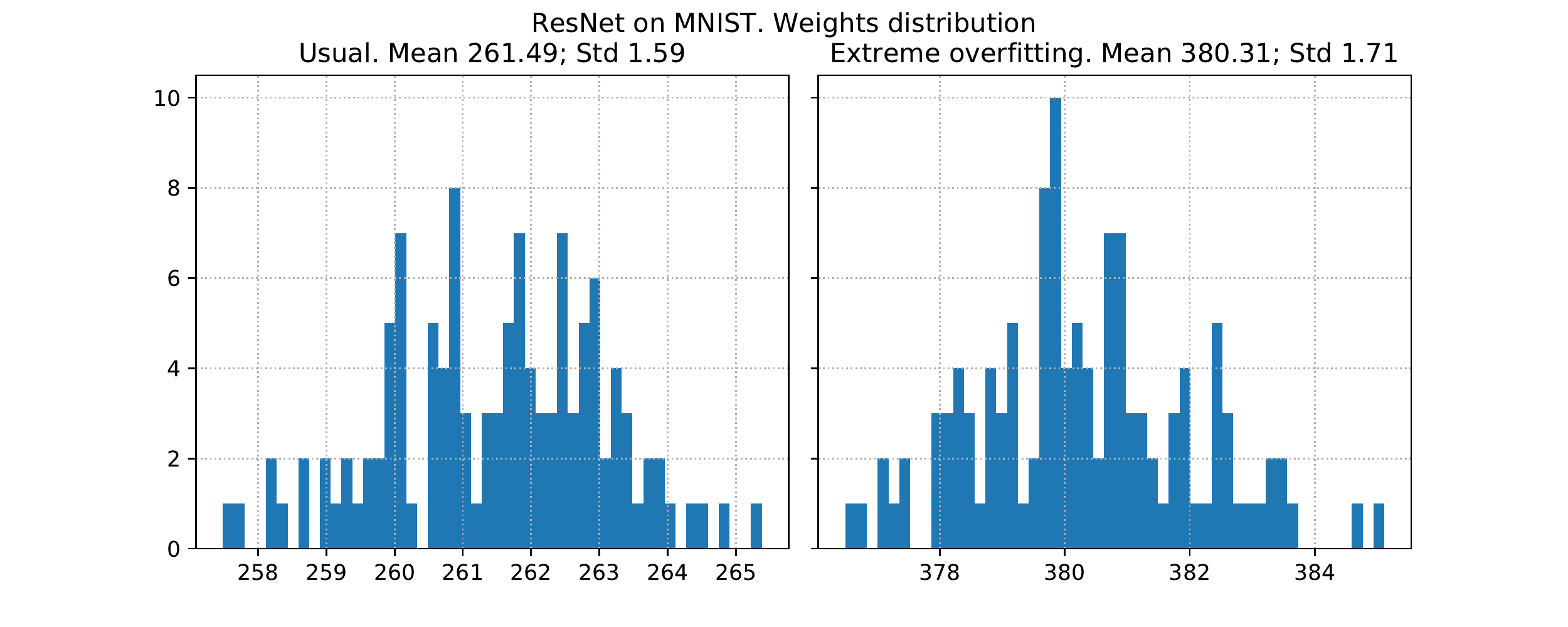}}
	\end{minipage}
	\hfill
	\begin{minipage}[h]{0.33\linewidth}
		\center{\includegraphics[width=1.1\linewidth]{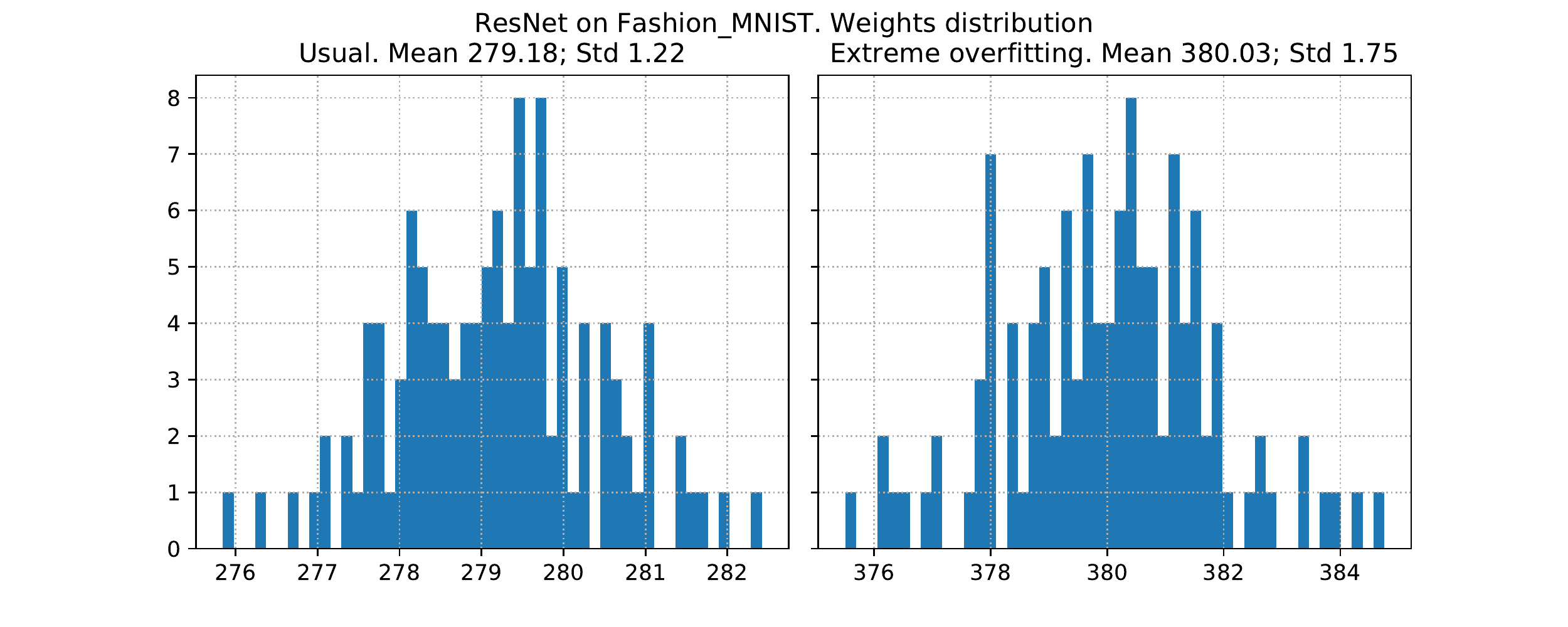}}
	\end{minipage}
	\hfill
	\begin{minipage}[h]{0.33\linewidth}
		\center{\includegraphics[width=1.1\linewidth]{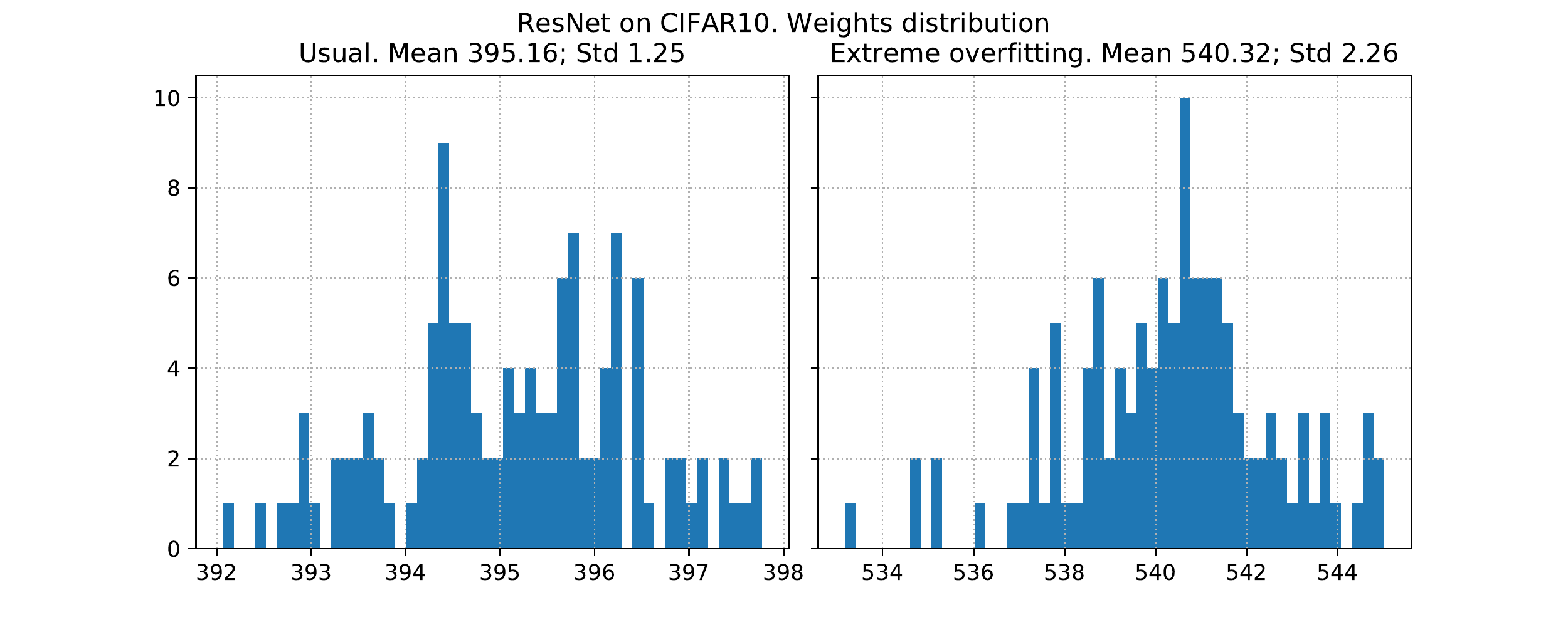}}
	\end{minipage}
	\begin{minipage}[h]{1\linewidth}
		\begin{tabular}{p{0.33\linewidth}p{0.33\linewidth}p{0.33\linewidth}}
			\centering a) MNIST & \centering b) Fashion MNIST & \centering c) CIFAR 10 \\
		\end{tabular}
	\end{minipage}
	\vspace*{-0.7cm}
	\caption{The distribution of weights in the process of usual training(left) and extreme overfitting (right). ResNet}
	\label{weights_ResNet}
\end{figure}

\section{Experimental setup}
\label{sec:exp_par}
\subsection{Neural Networks architectures and datasets}

\begin{figure}[H]
	\begin{minipage}[h]{0.4\linewidth}
		\center{\includegraphics[width=0.95\linewidth]{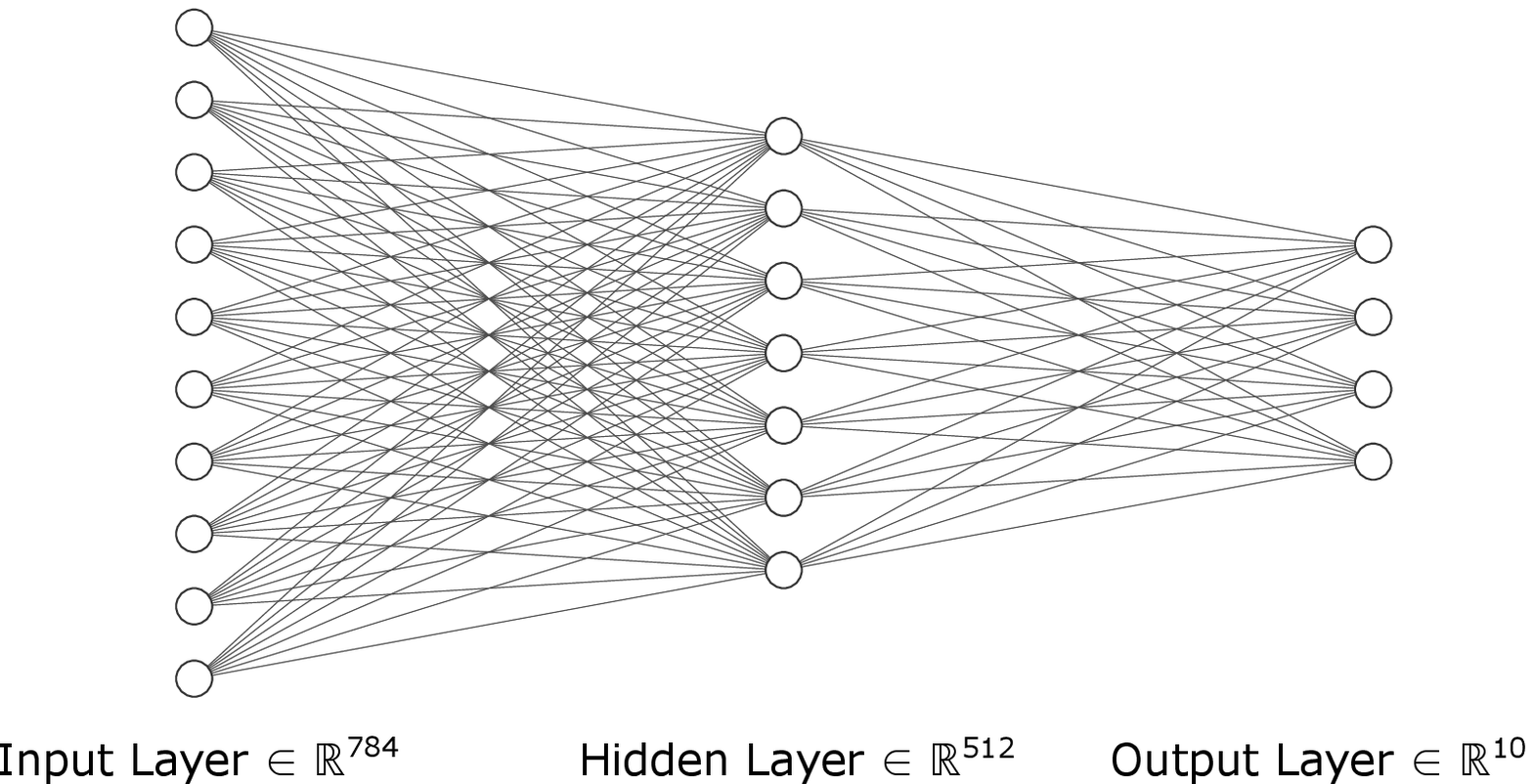}}
	\end{minipage}
	\hfill
	\begin{minipage}[h]{0.4\linewidth}
		\center{\includegraphics[width=0.95\linewidth]{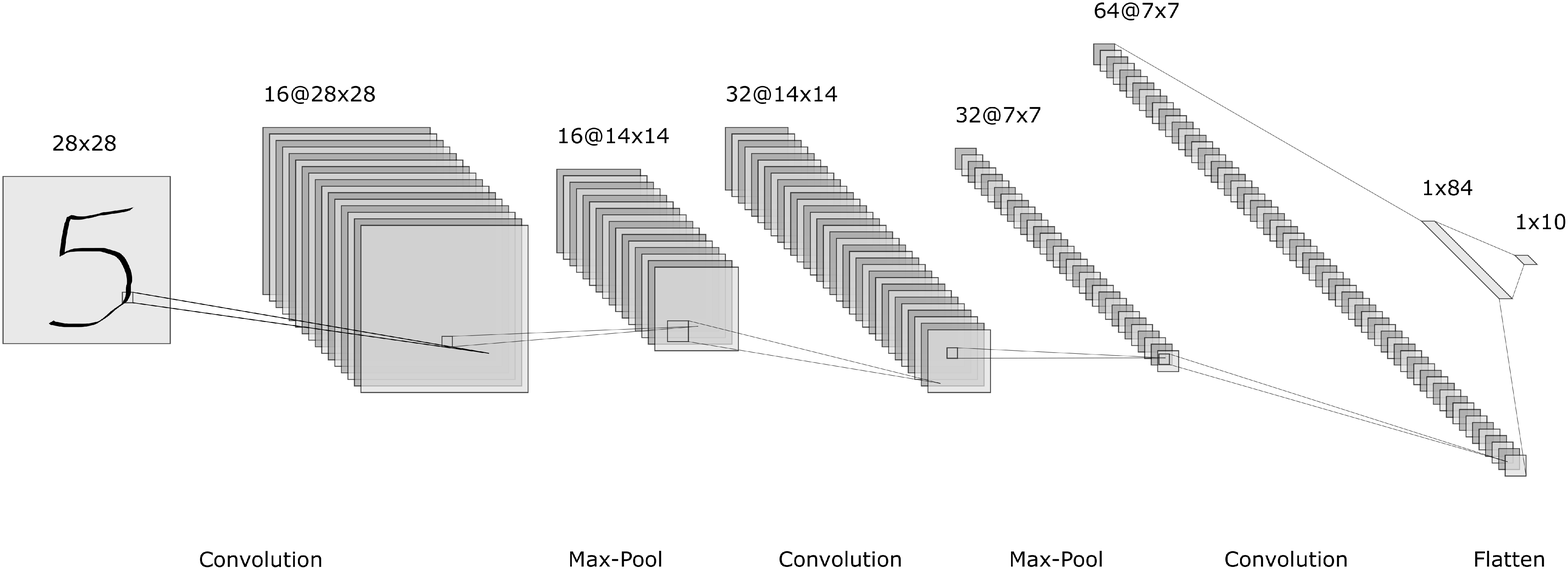}}
	\end{minipage}
	\hfill
	\begin{minipage}[h]{0.19\linewidth}
		\center{\includegraphics[width=1.2\linewidth]{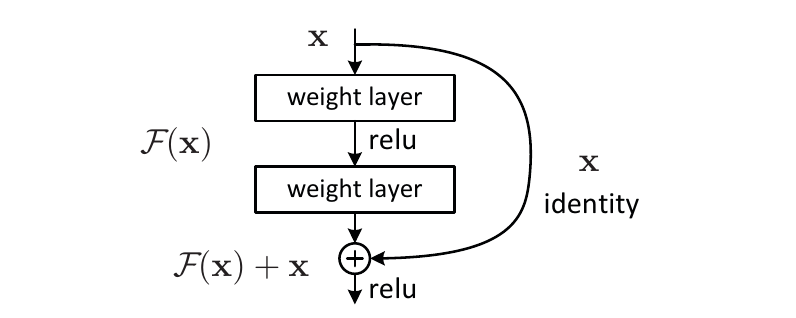}}
	\end{minipage}
	\begin{minipage}[h]{1\linewidth}
		\begin{tabular}{p{0.4\linewidth}p{0.4\linewidth}p{0.19\linewidth}}
			\centering a) & \centering b) & \centering c) \\
		\end{tabular}
	\end{minipage}
	\vspace*{-1cm}
	\caption{Studied DNN architectures: a) MLP, b) CNN, c) ResNet block\cite{he2016deep}.}
	\label{architectures}
\end{figure}

The following neural network architectures were considered: a fully connected neural network, a convolutional neural network that is inspired by LeNet \cite{le1989 handwritten} architecture, and also ResNet \cite{he2016deep}. Schemes of the considered architectures are shown in the figure \ref{architectures}.

For a fully connected neural network (MLP), we used three layers with nonlinear activations of the form $ ReLU (x) = \ max \{0, x \} $, the dimension of the input layer is equal to the dimension of the vectorized image for the classification supplied to the input, i.e. 784 for MNIST and Fashion MNIST and 3072 for CIFAR 10. The number of neurons in the hidden layer is 512. The number of neurons in the output layer corresponds to the number of classes, i.e. 10 for all data sets.
	
For the convolutional neural network (CNN), we used an architecture similar to LeNet \cite{le1989handwritten}. Three convolutional layers with a convolution core size of $ 5 \times 5 $ each contain 16, 32, and 64 filters, respectively. After the first two layers, the image size is reduced using the pooling method with a maximum function. The nonlinear activation function of the $ ReLU $ type is applied to the images obtained at the output of these layers. After the convolutional layers, the representation of the image is vectorized and fed to the input of a fully connected layer with the number of neurons 84, then $ ReLU $, then the last fully connected layer of $ softmax $ from among the neurons corresponding to the number of classes in the classification problem, i.e. 10 for all data sets.
	
For the ResNet neural network, the classical architecture presented in the \cite{he2016deep} article was used with 64 neurons in the penultimate fully connected layer and 10 neurons in the last fully connected layer. Also note that the architecture has been changed to apply it to data sets from black and white images MNIST and Fashion MNIST (it was originally created to work with color images in CIFAR 10 and ImageNet)

\begin{figure}[H]
	\begin{minipage}[h]{0.4\linewidth}
		\center{\includegraphics[width=0.95\linewidth]{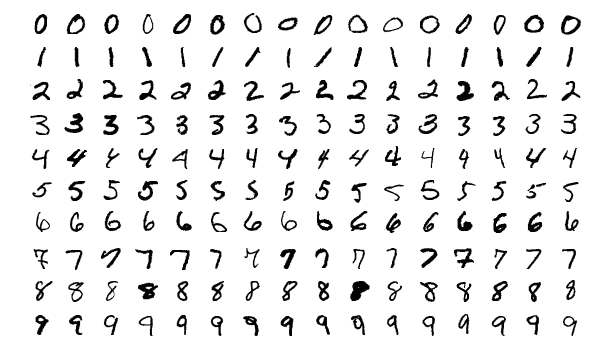}}
	\end{minipage}
	\hfill
	\begin{minipage}[h]{0.2\linewidth}
		\center{\includegraphics[width=0.95\linewidth]{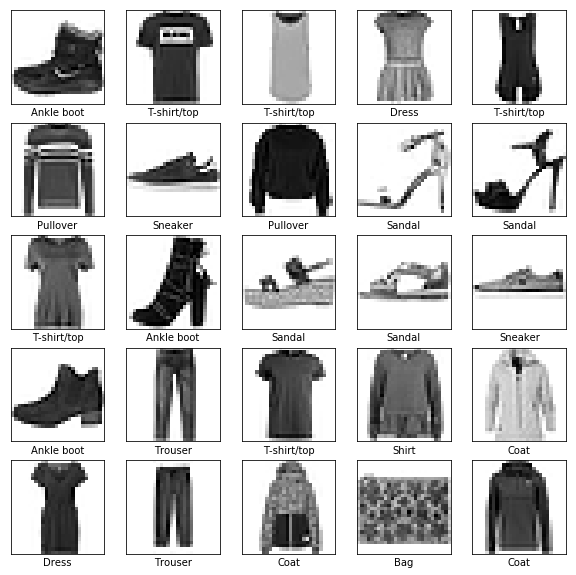}}
	\end{minipage}
	\hfill
	\begin{minipage}[h]{0.39\linewidth}
		\center{\includegraphics[width=0.95\linewidth]{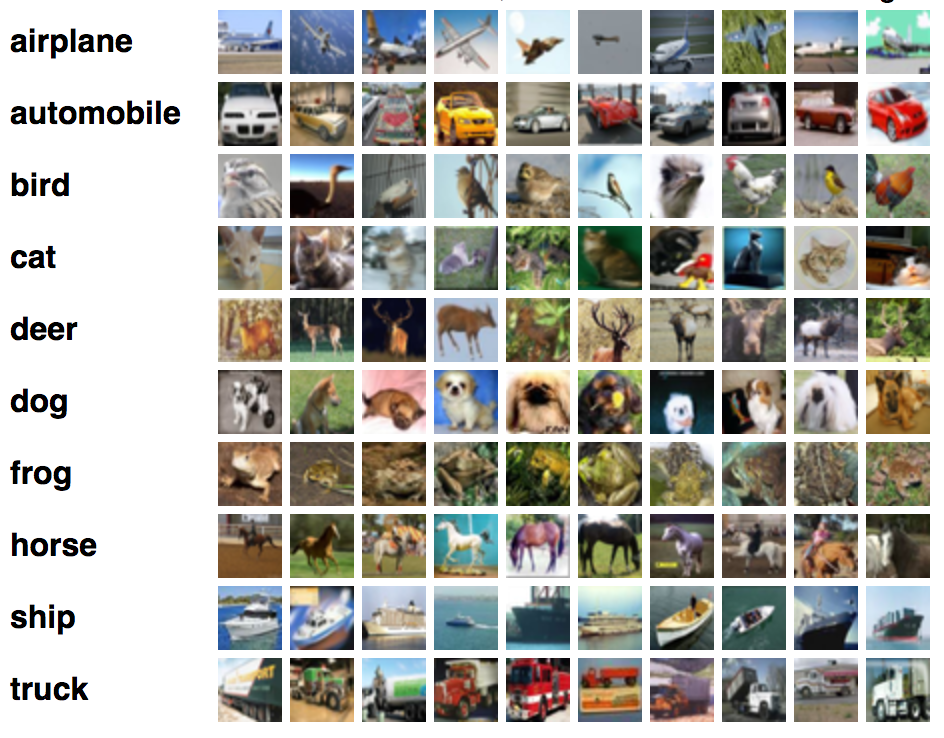}}
	\end{minipage}
	\begin{minipage}[h]{1\linewidth}
		\begin{tabular}{p{0.4\linewidth}p{0.2\linewidth}p{0.39\linewidth}}
			\centering a) & \centering b) & \centering c) \\
		\end{tabular}
	\end{minipage}
	\vspace*{-1cm}
	\caption{Images from datasets: a) MNIST, b) Fashion MNIST, c) CIFAR 10.}
	\label{dataset}
\end{figure}

The study was conducted using widely known in academic community datasets: MNIST \cite{lecun-mnisthandwrittetendigit-2010} Fashion MNIST \cite{xiao2017 / online}, CIFAR 10 \cite{krizhevsky2009learning}. Examples of images contained in these datasets for clarity, shown in the figure \ref{dataset}.

The MNIST dataset contains 60,000 grayscale images with sizes of $ 28 \times 28$ in a training sample and 10,000 images of the same size in a test sample. It includes images of handwritten digits, divided into 10 classes corresponding to Arabic numerals.

The Fashion MNIST dataset was created to test the machine learning algorithms created for the tasks of classifying images of a MNIST dataset in a new way, so it repeats the shape of the original dataset: 60,000 images of the same size in a training set and 10,000 in a test set. Images are divided into 10 classes: T-shirt/top, Trouser, Pullover, Dress, Coat, Sandal, Shirt, Sneaker, Bag, Ankle boot.

The CIFAR 10 dataset contains 50,000 color images of $ 32 \times 32 \times 3$ in a training sample and 10,000 images of the same size in a test sample. Images are divided into 10 classes: airplane, car, bird, cat, deer, dog, frog, horse, ship, truck.

\subsection{Software and hardware}

The numerical experiments were conducted on the basis of the computing cluster of the Skolkovo Institute of Science and Technology NVIDIA DGX-1 with 8 V100 video cards and an Intel (R) Xeon (R) E5-2698 v4 @ 2.20 GHz processor with 80 logical cores. The software is written in Python using the PyTorch \cite{paszke2017automatic} library. The experiments carried out allow a parallel launch, which allowed us to run different models on different video cards to speed up the learning process.

\subsection{Experimental methodology}

Neural networks were trained using stochastic first-order optimization algorithms. In particular, all the graphs presented in the work were obtained using the Adam \cite{kingma2014adam} optimization algorithm with a constant length step of 0.001 and a batch size of 128. The choice of these hyperparameters is due to the stability of the results in the study. In addition, experiments were conducted with the classical algorithm of stochastic gradient descent \cite{robbins1951stochastic}, however, the experimental results did not change qualitatively.

For each neural network architecture and for each dataset, we conducted 200 training sessions on a regular dataset and 200 training sessions on a corrupted dataset to obtain the typical weights that neural networks receive during usual training, as well as extreme overfitting points for further research. After each launch, the weights of the neural networks were initialized using the \texttt{torch.nn.init.xavier\_uniform} \cite{glorot2010understanding} method. The resulting weights were saved using the \texttt{torch.save} method for further analysis.

\section{Related Work}

Due to their extraordinary efficiency in practical tasks, neural networks became the focus of researches in a wide range of areas. Therefore, many works about different aspects of training neural networks were published. Here we will overview only small part of them, grouped by subject.

Vapnik \cite{vapnik1999overview} obtained results about bounding generalization error using $VC$-dimension concept. Some estimations on $VC$-dimension of neural networks were also obtained \cite{sontag1998vc}.

Zhang \cite{zhang2016understanding} showed through extensive systematic experiments, that modern DNN architectures can fit random labels.

Choromanska et al.\cite{choromanska2015loss} investigated loss surfaces of multilayer networks using spin-glass model. Under uniformity, independence and redundancy assumptions they showed, that there is an area, that band local minima with lowest critical values of random loss function. Also, they showed, that number of local minima outside such area decreases exponentially. Moreover, they numerically performed, that most local minima are equivalent in terms of test accuracy. 

Kawaguchi \cite{kawaguchi2016deep} enhanced statement, mentioned in \cite{choromanska2015loss}. He also considered linear networks with squared loss function and proved that every local minima of such loss surface is a global minima. Despite the usual assumption of similarity of nature of training and test data, we still can obtain sad points of neural networks.

\section{Conclusion}

The article proposes a method for obtaining \textit{points of extreme overfitting} - weights of modern deep neural networks, at which they demonstrate accuracy close to 100 \% to the training set, simultaneously with almost zero accuracy on the test sample. Such critical points of the loss function of the neural network, despite the widespread opinion that the vast majority of them have equally good generalizing abilities, have a large generalization error. In this paper, their properties were investigated, in particular, it was empirically shown that, on average, they are located much farther from the initialization weights than the points obtained during normal training. In addition, they do not interfere with stochastic gradient descent, since the initialization of such an algorithm for optimizing points leads to a critical point, which is not an extreme point of retraining.

The work contains systematic numerical experiments for modern models of deep neural networks, which are well suited for practical problems of image classification: fully connected networks, convolutional networks, and ResNet. For all data sets, we managed to get extreme overfitting points and study their properties. The presence of such points in neural networks is a good reason for further analytical study of the issue of training neural networks.

\end{document}